\documentclass[letterpaper, 10 pt, conference]{ieeeconf}  

\IEEEoverridecommandlockouts                              

\usepackage[utf8]{inputenc}
\usepackage{hyperref}
\usepackage{amsmath}
\usepackage{graphicx}
\usepackage{caption}
\usepackage{subcaption}
\usepackage{multirow}
\usepackage{bm}
\usepackage{rotating}
\usepackage{multirow}
\usepackage{hhline}
\usepackage{verbatim}
\usepackage{soul}
\usepackage{afterpage}
\usepackage{color}
\usepackage{latexsym}
\usepackage{dingbat}
\usepackage{pifont}
\usepackage{amssymb}
\usepackage{amsmath}

\usepackage{pdflscape}
\usepackage{enumerate}
\let\labelindent\relax
\usepackage[inline]{enumitem}
\usepackage{xspace} 
\usepackage{amsmath}
\usepackage[ruled]{algorithm2e}
\usepackage{algorithmicx}
\usepackage{float}
\usepackage[table]{xcolor}
\usepackage[normalem]{ulem}
\usepackage{makecell}
\usepackage[american, europeanresistors]{circuitikz}

\usepackage{pifont}
\newcommand{\dq}{{\boldsymbol {\dot q}}}

\newcommand{\bx}{{\boldsymbol x}}
\newcommand{\dx}{{\boldsymbol {\dot x}}}
\newcommand{\ddx}{{\boldsymbol {\ddot x}}}

\newcommand{\bC}{{\boldsymbol C}}

\newcommand{\bF}{{\boldsymbol F}}

\newcommand{\bJ}{{\boldsymbol J}}
\newcommand{\bJT}{{\boldsymbol{J}^T}}

\newcommand{\btau}{{\boldsymbol \tau}}

\newcommand{\bG}{{\boldsymbol G}}

\newcommand{\bV}{{\boldsymbol V}}

\newcommand{\bLambda}{{\boldsymbol \Lambda}}

\newcommand{\bsq}{{\boldsymbol q}}

\SetKwData{Left}{left}\SetKwData{This}{this}\SetKwData{Up}{up}
\SetKwFunction{Union}{Union}\SetKwFunction{FindCompress}{FindCompress}
\SetKwInOut{Input}{input}\SetKwInOut{Output}{output}

\newlength\kwinlength

\newlength\kwoutlength

\usepackage[per=slash]{siunitx}
\usepackage{comment}

\definecolor{OliveGreen}{RGB}{0,200,25}
\newcommand{\red}[1]{\textcolor{red}{#1}}

\newcommand{\darkgreen}[1]{\textcolor{OliveGreen}{#1}}

\newcommand{\eg}{e.\,g.\ }

\newcommand{\armarVI}{\mbox{ARMAR-6}\xspace}

\newcommand{\ackSecondhands}{The research leading to these results has received funding from the European Union’s Horizon 2020 Research and Innovation programme under grant agreement No 643950 (SecondHands).}

\newcommand{\replaced}[2]{\red{\ifmmode\text{\sout{\ensuremath{#1}}}\else\sout{#1}\fi}\darkgreen{#2}}
\newcommand{\removed}[1]{\red{\ifmmode\text{\sout{\ensuremath{#1}}}\else\sout{#1}\fi}}

\newcommand{\removedfootnote}[1]{\footnote{\removed{#1}}}
\newcommand{\removedsubsection}[1]{\subsection{\texorpdfstring{\removed{#1}}{#1}}}

\usepackage[per=slash]{siunitx}
\usepackage{amsbsy}

\newcommand{\mSig}{\pmb{\Sigma}}

\newcommand{\mmu}{\pmb{\mu}}
\newcommand{\Normal}{\mathcal{N}}

\newcommand{\mV}{\pmb{V}}
\newcommand{\mF}{\pmb{F}}
\newcommand{\mTh}{\pmb{\Theta}}

\title{Learning Compliance Adaptation in Contact-Rich Manipulation}
\author{Jianfeng Gao, You~Zhou and~Tamim~Asfour
\thanks{The authors are with the Institute for Anthropomatics and Robotics, Karlsruhe Institute of Technology, Karlsruhe, Germany. {\tt \{jianfeng.gao, you.zhou, asfour\}@kit.edu}. }
\thanks{
The first two authors contributed equally to this work. Their names are alphabetically ordered. }
\thanks{\ackSecondhands}
}

\begin{document}
\inputencoding{utf8}

\maketitle
\thispagestyle{empty}
\pagestyle{empty}

\begin{abstract}
Compliant robot behavior is crucial for the realization of
contact-rich manipulation tasks. In such tasks, it is
important to ensure a high stiffness and force tracking accuracy
during normal task execution as well as rapid adaptation
and complaint behavior to react to abnormal situations and
changes. 
In this paper, we propose a novel approach for learning
predictive models of force profiles required for
contact-rich tasks. Such models allow detecting unexpected
situations and	facilitates better adaptive control. The
approach combines an anomaly detection based on
Bidirectional Gated Recurrent Units (Bi-GRU)  and an
adaptive force/impedance controller. 
We evaluated the approach in simulated and real-world
experiments on a humanoid robot.The results show that the
approach allow simultaneous high tracking accuracy of
desired motions and force profile as well as the adaptation
to force perturbations due to physical human interaction.
\end{abstract}

\section{Introduction} \label{sec:introduction}

In human-centered applications, robots should be able to perform a wide variety of the tasks in a compliant way to meet the safety requirement.  To assist human, the robot usually manipulates an object in a certain way or works with a specific tool for a task. This could be addressed as motion and force control problem in a contact-rich environment. 
For example, the robot cleans an arbitrary unknown surface with a sponge. The surface is static or dynamically varying and the human might also interrupt the cleaning task. 
We should also consider the situation where the robot might not always see through vision system what is going to happen in the dynamic environment. Thus, creating compliant controllers based only on the motion and force feedback is necessary. 


Some tasks require a large force to be exerted on the objects or the tools, such as cleaning heavily polluted surface. 
The robot should keep high stiffness to ensure the quality of the task execution. 
However, since the dynamic model for a contact-rich manipulation is difficult to obtain and the robot needs \eg to react to sudden collision in a human-centered environment.
A conventional model-based control system with high stiffness might result in dangerous actions and thus can usually not be applied in such scenarios. 

By observing human wiping and interacting with others, we found some major abnormal events, for example, another person interrupts the subject by blocking the trajectory, the subject collides with an object or another person or suddenly loses contact when wiping out of the surface. 
These will happen to the robot as well. 
Therefore, the robot must learn how to distinguish between normal and abnormal situations for a certain task based on motion and force feedback, as we do. 
This can be achieved by training a predictive model as an anomaly detector based on the data collected during normal executions of the tasks and using it in the future execution.

Proper adaptation law is also needed to modify the compliance behavior in different situations so that the robot could be super compliant and thus maintain safe.
When the situation goes back to normal, the control strategies in the normal situation will be resumed to continue the task with higher stiffness.


\begin{figure}[t]
		\centering
		\includegraphics[width=0.9\linewidth]{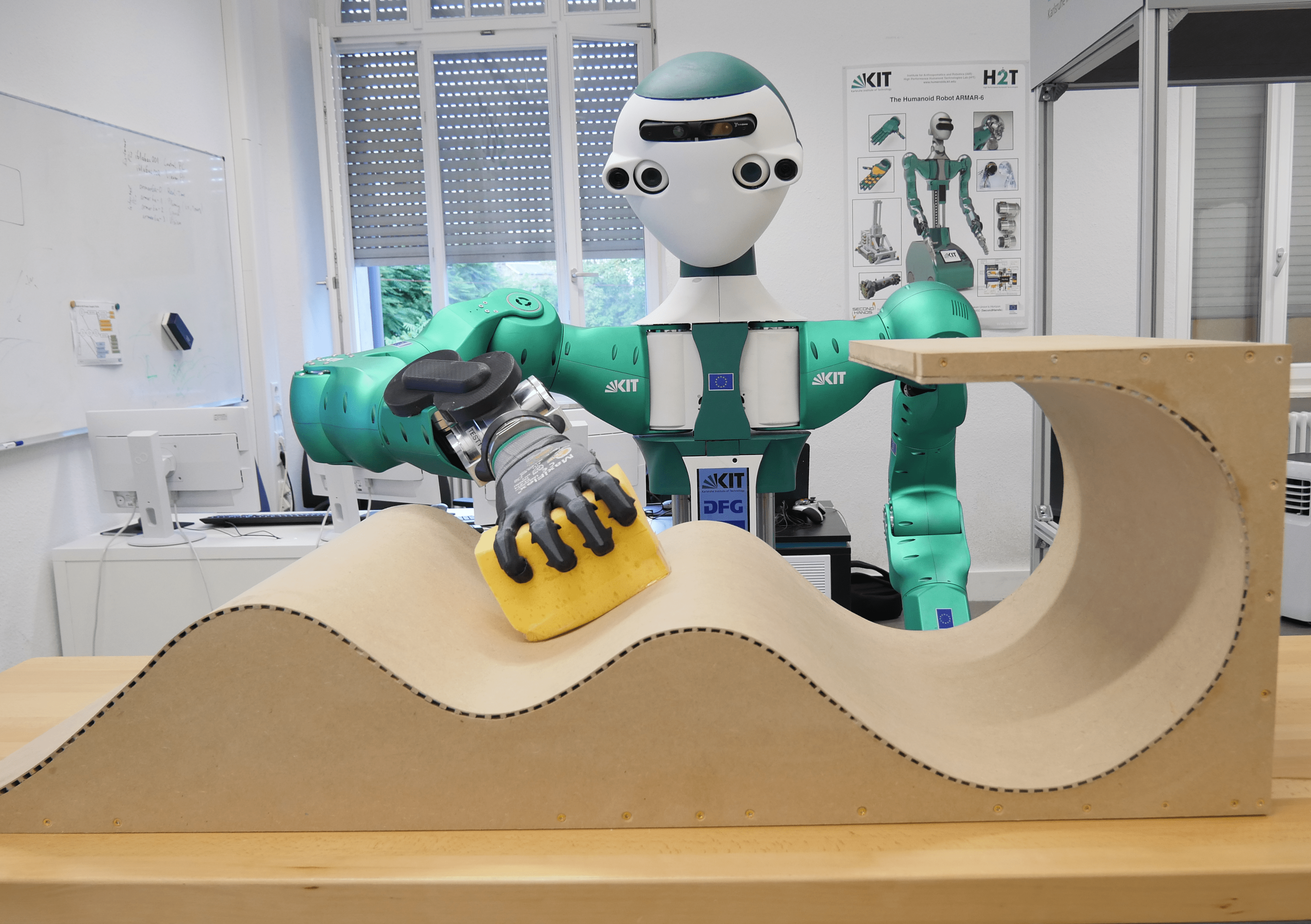}
		\caption{\armarVI cleans an arbitrary surface. }
		\label{fig:clean_arbitrary_surface}
\end{figure}

\section{Related Works} \label{sec:related_works}

In order to realize both high tracking accuracy and compliant behavior, the authors in~\cite{Denisa2015} proposed to learn a mapping from the task parameters to the position and torque trajectories. 
The position trajectories were obtained by human demonstrations. While the torque trajectories were recorded with a high stiffness tracking controller. After learning the torque trajectory. 
A torque based compliant controller can track the target position trajectory accurately with compliant behavior. 
However, with the proposed method, the robot can only stop and restart from the beginning when encountering the external disturbance, because the learned torque controller does not know how to recover from the disturbance.

In the active interaction control field, there are multiple approaches to achieve fast, stable and coupled force-motion control, such as indirect and direct force control~\cite{Villani2008}. The former approach, e.g. hybrid impedance control~\cite{Hogan1984}~\cite{Anderson1988}, shows stable compliant behavior by properly setting the stiffness of the robot and a biased trajectory w.r.t the modeled surface. However, HRI usually involves unmodeled environment and unexpected human interactions, thus, requires time-varying compliance behavior. 

One of the aspects of direct force control is the explicit force control, which makes the robot to be an effort-source. The force and torque feedback signals are used to calculate the force control command directly, without using inner-loop motion control. 
This feature makes it respond faster to the sudden change of the environment. However, it's also more sensitive to impact or losing contact and more vulnerable to instability factors in the real-time control system. 
To observe its stability, the authors in ~\cite{Balachandran2017}~\cite{Jorda2017} proposed several passivity-observer based approaches to adapt the gain of the direct force control strategy. Another example is the task-energy tank based unified force-impedance control~\cite{Schindlbeck2015a}, which shows safe adaptation when loose-contact is detected. 
However, designing passivity-observer for complex interaction tasks is not easy. 


An another aspect of direct force control is to add an inner motion control loop. The outer control loop, e.g. explicit force control, sends a position or velocity command to the inner loop. 
A compliant and asymptotically stable behavior could be achieved by assigning a critically or over-damped impedance to the motion controller. 
An admittance-coupled DMP framework is proposed in~\cite{6748918} and later is combined with passivity analysis and iterative learning control (ILC)~\cite{Shahriari} to achieve online adaptation to changing surface. However, ILC based adaptation is quite time-consuming, hence does not fit the contact-rich unmodeled environment.
%

\section{Basic Knowledges}
\subsection{Motion Representation}
\label{sec:motion_representation}

\begin{figure*}[ht]
  \centering
    \includegraphics[width=\textwidth]{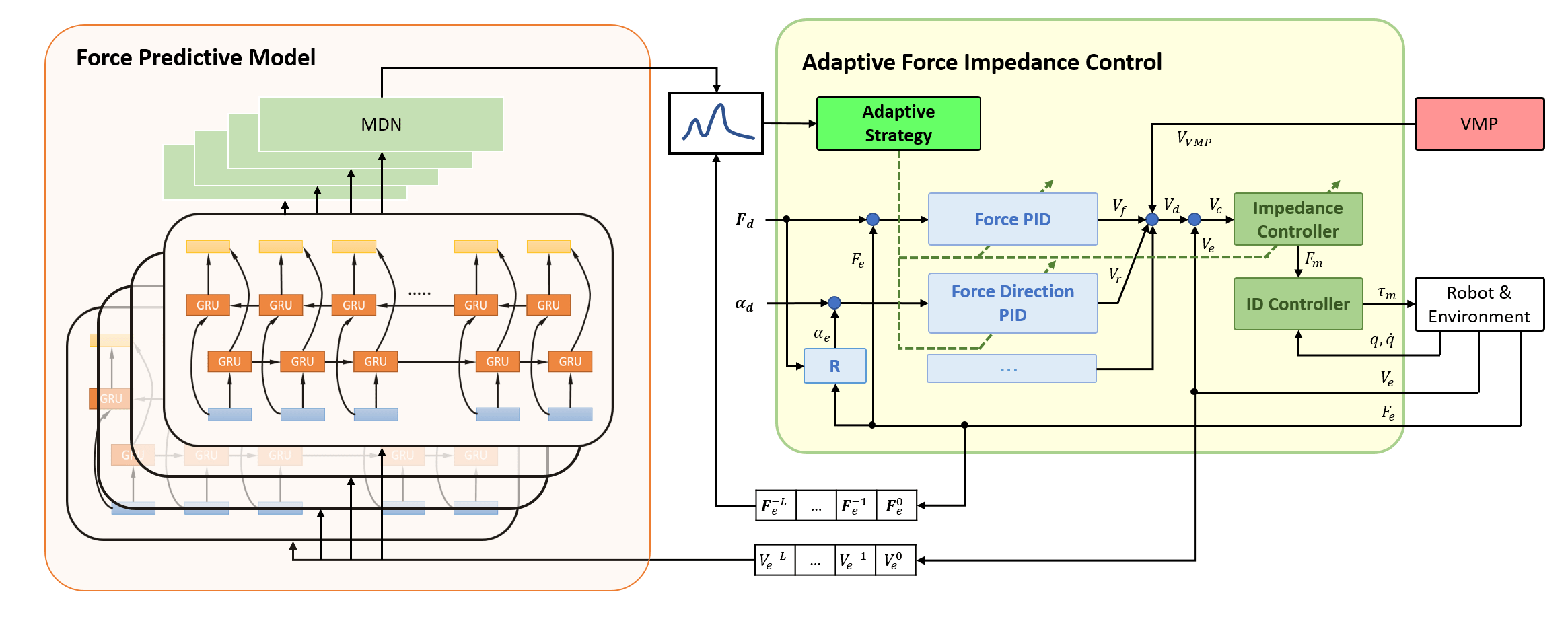} 
  \caption{The proposed system consists of two different parts: adaptive force impedance control and force predictive models. \textbf{Right:} The control framework indicated by the light yellow block contains multiple adaptive controllers, which tracks the target force and direction in the contact-rich manipulation. 
Velocity sources include VMP for motion tracking and PID force controllers. Force command is calculated by the impedance controller and finally mapped to the joint torque commands by the inverse dynamic controller. 
\textbf{Left:} Several bi-directional GRU models are used as the force predictive models to predict the mixture distribution of the force profiles. 
The learned distribution is then used to evaluate the current force profile regarding the contact-rich manipulations to check whether it is abnormal or not. 
The results further affect the adaptive strategy to decide which control mode takes over. }
 \label{fig:control_diagram}
\end{figure*}

In order to represent the learned motion from human demonstrations, we use Via-points Movement Primitive (VMP) presented in our previous work~\cite{Zhou2019}. VMP consists of two parts: elementary trajectory and shape modulation:
\begin{equation}\label{eq:vmp}
y(x) = h(x) + f(x),
\end{equation}
where the elementary trajectory directly connects the start and end points. While the shape modulation encodes the shape of the trajectory. In the wiping experiment, the elementary trajectory is the position of the anchor point, which can be changed to translate the whole wiping motion. 
The shape modulation encodes the wiping pattern. Both parts of VMP are differentiable, which allows VMP to provide both target position and velocity. 
The target velocity of VMP provides the velocity source for the force impedance controller mentioned in~\autoref{subsec:imp-control} (see~\autoref{fig:control_diagram}). 

\subsection{Impedance Controller}\label{subsec:imp-control}


\begin{figure}[t]
	\centering
	\includegraphics[width=\linewidth]{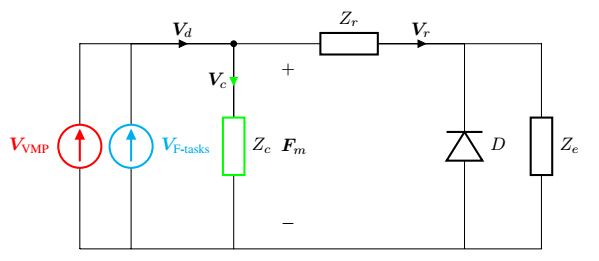} 
	\caption{Impedance Control Framework. Desired motion is encoded as a VMP velocity source and force PID controllers are additional velocity sources for tracking of target force profile and force direction regulation. The environment is treated as an effort source.}
	\label{fig:flow_source}
\end{figure}

Since the explicit force control with an inner motion control loop shows good stability and extensibility feature, we choose this framework as the base for our controllers. 
Within this framework, the environment is treated as an effort-source and the controller makes the robot as a velocity-source as shown in Fig~\ref{fig:flow_source}. Two major velocity sources are used. One is for motion control, which is encoded in VMP ($\bV_{\text{VMP}}$). 
The other one is the PID-based force controllers ($\bV_{\text{F-tasks}}$), which is used to track a force profile or regulate the desired force direction. 
The total desired velocity, the control-flow and the velocity of the robot are represented as $\bV_d, \bV_c, \bV_r$ respectively. 
The impedance of the motion controller, the robot and the environment is denoted as $Z_c, Z_r$ and $Z_e$. 
The diode means that we only deal with uni-literal contact. 


\section{Adaptive Control with Force Predictive Models}

\subsection{Adaptive Force Impedance Control}\label{subsec:force_impedance}
%

The task space dynamic equation of the robot is:
\begin{equation}
\bLambda \ddx  + \bC (\bx, \dx) \dx + \bG (\bx) = \bF_m + \bF_e,
\end{equation} 
where $\bLambda, \bC, \bG$ are task-space inertia matrix, Coriolis matrix and gravititional force respectively, and $\bF_m, \bF_e \in \mathbb{R}^6$ are the actuation force and contact force applied on the robot. 

We design the adaptive compliant controller by extending the velocity source for force control based on the impedance control framework as described in the last section and embedding an adaptive law based on the anomaly detector (see section \ref{sec:anomaly_detection}). 

The velocity source for motion control is based on our previous work about VMP, 
which provide the velocity reference $\bV_{VMP}$. 
To track the target force profile $\bF_d$, a PID controller is added to the second velocity source. In frequence domain, the controller is expressed by 
\begin{equation}
\bV_f(s) = (\bF_d(s) - \bF_e(s)) ( K_{pf} + \frac{K_{if}}{s} + sK_{df} ),
\end{equation}
where the $K_p, K_i, K_d$ are PID gains.
In addition, PID controllers to regulate the desired force direction $\alpha_d$ or to track a torque profile $\btau_d$ are added in the same manner as PID force controller.
And the control flows are denoted as $\bV_r(s), \bV_t(s)$ respectively.
The total velocity source is the combination of all these controllers described above,
\begin{equation}
\bV_d(s) = \bV_{VMP}(s) + \bV_f(s) + \bV_r(s) + \bV_t(s).
\end{equation}

The task space control force applied to the robot is
\begin{equation}
\bF_m = (\bV_d(s) - \bV_e(s)) (D_c + \frac{K_c}{s} + sM_c).
\end{equation}
The torque control command is computed by
\begin{equation}
\btau_m = \bJT(\bF_m + \bC (\bx, \dx) \dx + \bG (\bx)) + \mathcal{N}(\bsq, \dq),
\end{equation}
where $\bJ$ is the Jacobian matrix and $\mathcal{N}(\bsq, \dq)$ is the null-space controller.

In order to guarantee the tracking accuracy for both VMP encoded trajectory and target force profile, we consider obtaining the velocity sources with high stiffness impedance controller and high gain PID controllers. 
The impedance parameters $K_c, D_c, M_c$, and PID parameters, however, can change with two different modes: adaptive mode and recovery mode. With $K_{p}$ used to represent all these adaptable parameters for simplification, in the adaptive mode, we have
\begin{equation}\label{eq:adaptive_mode}
K_{p}(t) = max (K_{p}(t_0) - \alpha_{p} ( t - t_0), 0),
\end{equation}
and in the recovery mode,
\begin{equation}
K_{p}(t) = min (K_{p}(t_0) + \beta_{p} ( t- t_0), K_{p,max}),
\end{equation}
where $t_0$ is the timestamp when the mode is switched and  $t$ is the current timestamp. $K_{p,max}$ is the corresponding maximum value 

During the task execution, one of these two modes takes over. The adaptive mode modifies the robot to an extreme compliant status and the recovery mode allows the robot to continue the original task. 
In this work, the high stiffness $K_{p,max}$ and adaptive parameters $\alpha$ and $\beta$ are empirically determined. They are so determined to allow a rapid drop from $K_{p,max}$ to $0$ in less than one second in the adaptive mode and recover relatively slower to keep safe and avoid control instability.

\subsection{Learning Force Predictive Models } \label{sec:anomaly_detection}

In order to be able to know when to switch the mode, we consider creating an anomaly detector which can detect the abnormal force observed by the force sensor. Once the anomaly in the current force profile is identified, the adaptive mode is activated to adapt to the external disturbance such as human interruption or the collision with the object. 
Anomaly detection can take the normal force profile into consideration and allow compliance for the contact-rich tasks. Similar to~\cite{Denisa2015}, we assume that there exist a set of high stiffness controllers with appropriate parameters mentioned before that are stable and perform well in the task. 
And we allow that the high stiffness controllers run at least once to execute the task without the external disturbances. For example, in the periodic task such as wiping, it is not difficult to find one period where no disturbance occurs.

As an example, during the robotic wiping, the friction between the wiping tool and surface is necessary for the task. 
In this case, in order to guarantee the motion on the wiping surface, an impedance controller with high stiffness is required, which does not allow human interaction or can cause problems when the robot's end-effector collides with some obstacles. 
With an anomaly detector, we can identify the abnormal force and switch to the former mentioned adaptive control mode.

The straightforward anomaly detection is mainly based on the one class classification, where a classifier is trained only on the non-anomalous data. 
Some other methods are based on the unsupervised learning models such as an autoencoder, which can learn the underlying structure of the data. The reconstruction error of a well-trained model indicates an anomaly. 
For the time-series data like in our case, a predictive model can be constructed and trained on the non-anomalous sequences. 
The prediction error of this trained predictive model is evaluated to determine whether the new data is abnormal or not. 

For many contact-rich tasks, there is a tight relationship between the motion of the robot's end-effector and the force exerted on it:
\begin{equation}
F_{t+1} = f(v_t, h_t),
\end{equation}
where $F_{t+1}$ is the force at the next time step and $v_t$ is the current velocity. In the wiping task, both are represented in the local frame to have more generalizability. $h_t$ is the hidden state related to some temporal features. 
In the wiping experiment, this relationship might encode the shape of the wiping surface.  A bidirectional gated recurrent unit (Bi-GRU) is created as the predictive model, which is connected with a mixture density network (MDN) (\cite{Bishop1994}) to output a Gaussian mixture distribution of the force profile based on the current velocity profiles and hidden states.

During the training, we minimize the negative-log-likelihood (NLL) of the observed target force profile such that
\begin{equation}\label{eq:nll}
\begin{split}
l_{NLL}(\mTh) = &-\sum_{i=1}^{N} log \bigg( \sum_{k=1}^K \pi_k(\mV; \mTh) 
\\
& \Normal \Big( \mF ; \mmu(\mV; \mTh), \mSig(\mV; \mTh) \Big) \bigg),
\end{split}
\end{equation}
where $\mTh$ is the parameters of the whole network and $\mmu(\cdot)$ and $\mSig(\cdot)$ are the result functions for both mean and covariance. $\mV$ and $\mF$ are sequential velocity and force profile which are collected by the sliding window with fixed sequence length $L$ (which is $20$ in our experiments).

\subsection{Force Predictive Model based Adpative Strategy}

\label{subsec:adaptive_strategy}
\begin{figure}[t]
		\centering
		\includegraphics[width=\columnwidth]{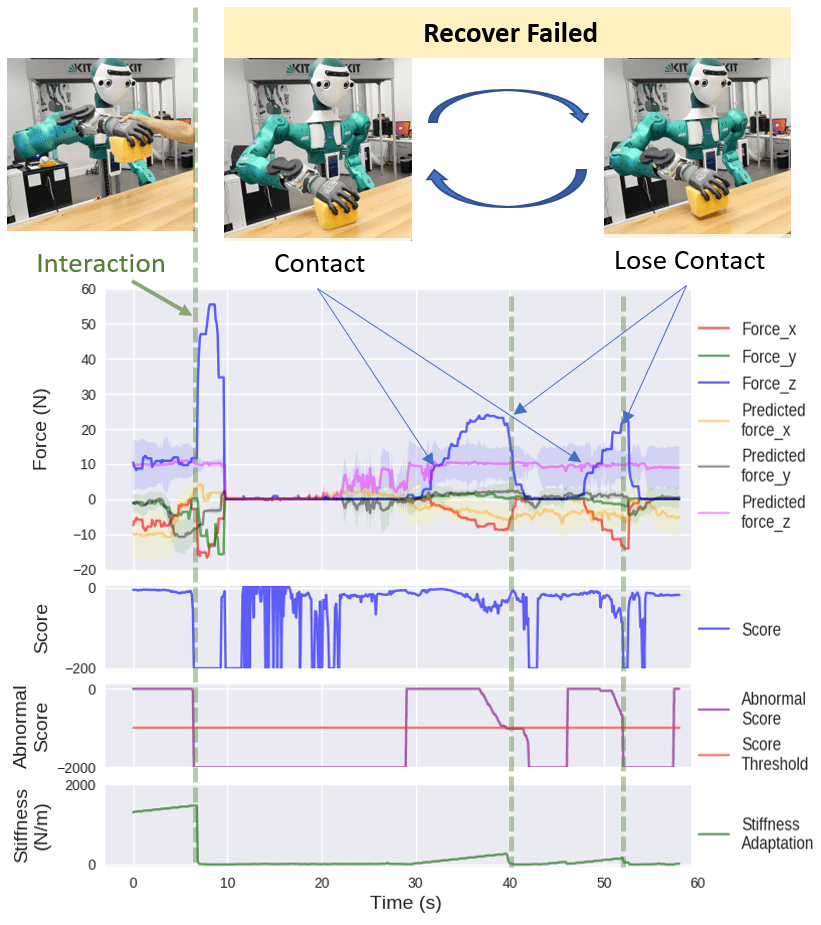}
		\caption{A single predictive model can cause the problem in the recovery mode. The robot might be stuck and cannot resume to the original tasks. }
		\label{fig:anomaly_plotting_single}
\end{figure}

In the task execution, the last $L$ steps velocities and forces are collected in two separate queues. The velocity queue is sent to the trained Bi-GRU model, which outputs a Gaussian mixture distribution of the force profile. With this distribution, we can calculate the probability of the collected force queue. The logarithm of this probability is called score. And its integral along the time is called abnormal score. We use integral to avoid reactions on some sudden and short disturbances which should not affect the robot task execution. In~\autoref{fig:anomaly_plotting}, we show that this score can drop very fast because of the logarithm function. Once the abnormal score is low, the control mode is switched to the adaptive mode. The simple linear adaptation of the control parameters is triggered (see~\autoref{eq:adaptive_mode})

We don't expect to have enough data to learn a well generalizable model, hence, allow overfitting to simplify the training process. This is suitable for many repeatable tasks such as wiping, where each period can have similar force profiles if the environment does not change a lot. However, for a changing enviornment, one single Bi-GRU might not be enough to generalize to different situations, especially when it might overfit the training data. An overfitted model will annotate every situation which is slightly different from the original training data as low probability, thus, abnormal event, where the control parameters are almost all zero according to our linear adaptation strategy. Hence, we consider to add more than one predictive models.  

For a specific task, the overall process is as follows:
\begin{enumerate}
\item At the beginning, no predictive models exist. We let the robot execute the task with a high stiffness controller and collect data for training the first predictive model;
\item Once we have at least one predictive model, we let the robot execute the task with the adaptive controller based on the control diagram in~\autoref{fig:control_diagram};
\item If the behavior of the robot is not as expected, such as stopping during the recovery control mode or during task execution without any external interference,  we collect those data to train an another predictive model. All predictive models work together to output a hierarchical mixture distribution according to the sum rule of the probability. 
\end{enumerate}
By iterating between step $2$ and $3$, we can finally end up with a combined predictive model which takes all situations that might occur during task execution and recovery from the disturbance into the considerations.

For the wiping task as an example, the first predictive model is trained in the normal situation, where wiping is conducted with high stiffness controllers and without any disturbance. However, with only this model, the robot can be stuck during the recovery mode, which is shown in~\autoref{fig:anomaly_plotting_single}. This can be avoided by learning the second predictive model that is trained with the collected force profile shown in~\autoref{fig:anomaly_plotting_single}. With the help of this predictive model, the robot can go through the recovery mode and resume the high stiffness controller. This strategy can be used also for the change of task parameters. For example, the friction of the table has a huge change, in which case the first learned predicitive model always prevents the robot from the wiping task.

\section{Experiments and Evaluations} \label{sec:evaluation}

To evaluate our method, we mainly focus on the wiping task in both simulator and real enviornment.

\subsection{Result from Simulation}

\begin{figure}[t]
	\centering
	\begin{subfigure}{0.15\textwidth}
		\centering
		\includegraphics[width=\textwidth]{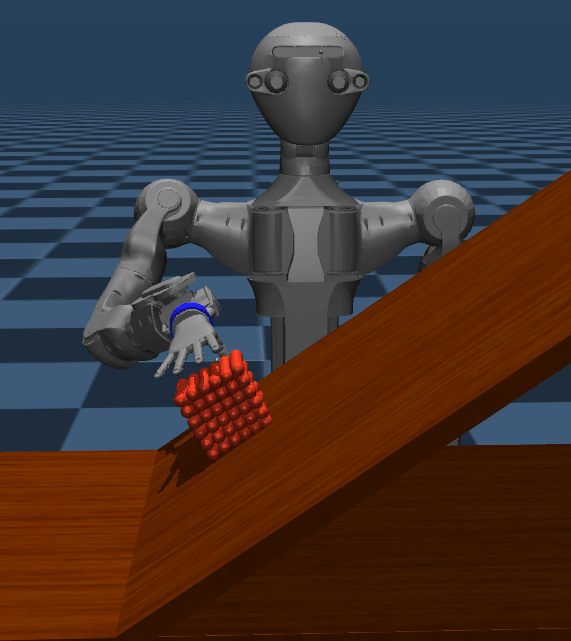} 
		\caption{Slope}
		\label{fig:mujoco_slope}
	\end{subfigure}
	\hspace{0.00mm}
	\begin{subfigure}{0.15\textwidth}
		\centering
		\includegraphics[width=\textwidth]{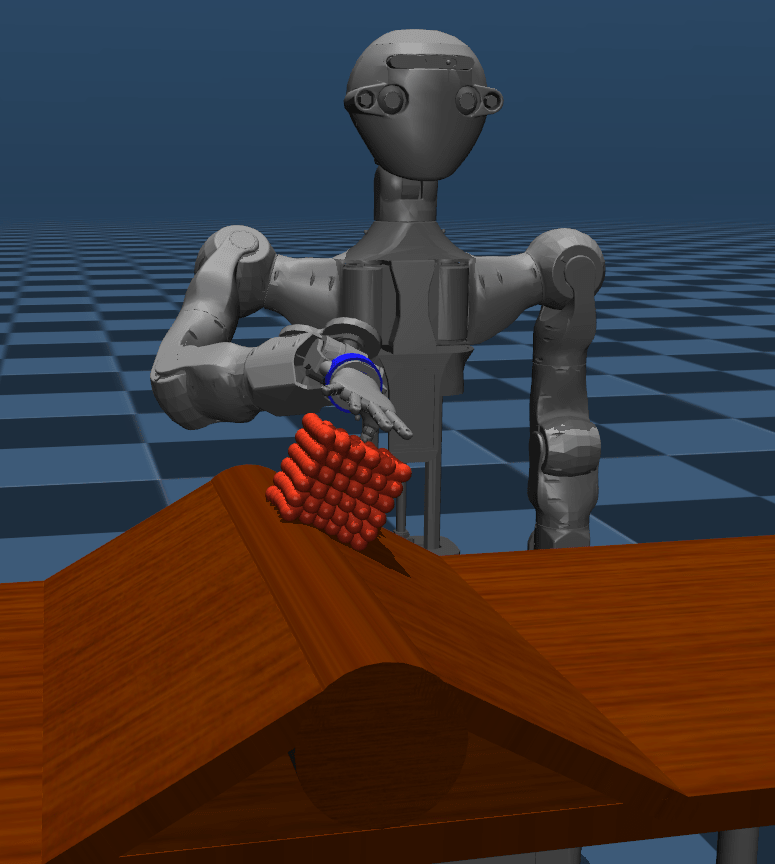} 
		\caption{Threshold}
		\label{fig:mujoco_stage}
	\end{subfigure}
	\hspace{0.00mm}
	\begin{subfigure}{0.15\textwidth}
		\centering
		\includegraphics[width=\textwidth]{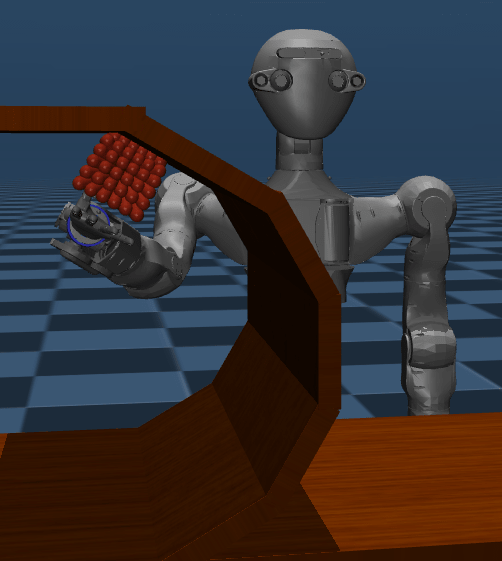} 
		\caption{Half Circle}
		\label{fig:mujoco_half_circle}
	\end{subfigure}
	\label{fig:mujoco}
	\caption{Experiments in Mujoco show the force based orientation control of our wiping controller.}
\end{figure}

In the simulator as shown in Fig~\ref{fig:mujoco_slope}, we setup surfaces with different slope and friction coefficient. To automatically detect the friction coefficient, which is necessary to extract current force direction, we add a simple online linear regression model to our controller. The parameters are tuned such that the robot can accurately track arbitrary motion and force profile. Then the data for training anomaly detector is collected when the robot wipes the surface with high stiffness.

\begin{figure}[t]
	\centering
	\includegraphics[width=0.5\textwidth]{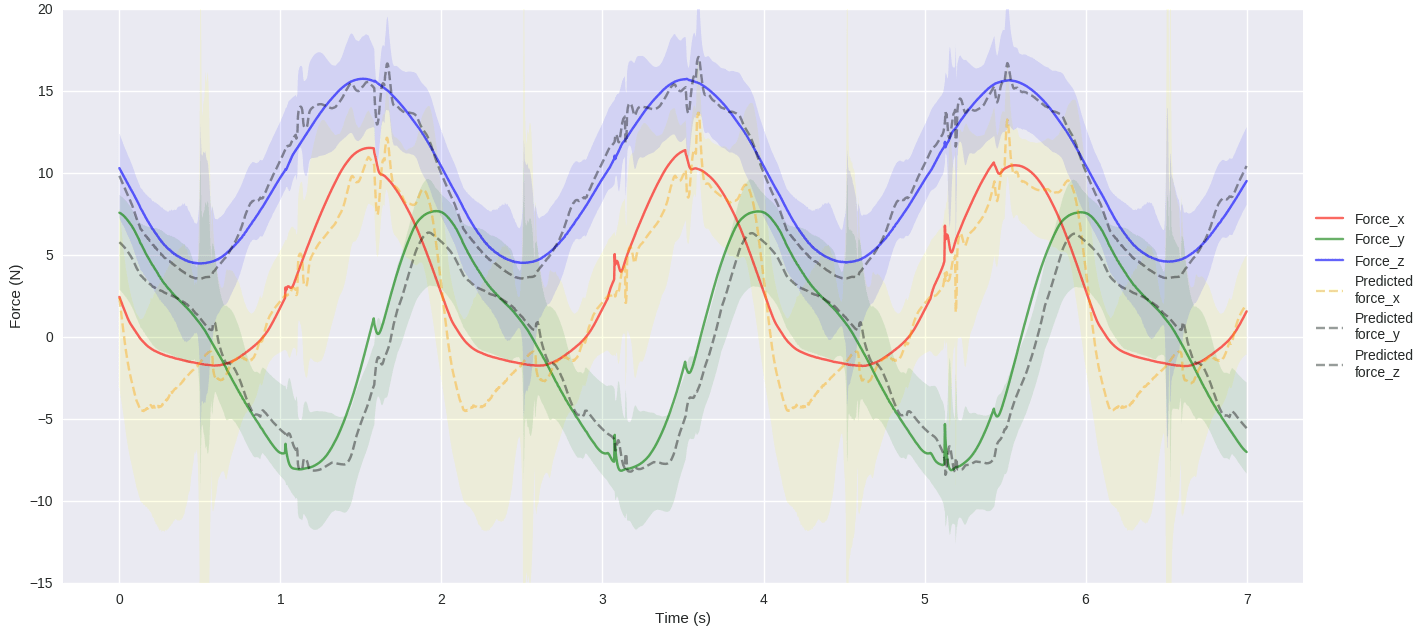} 
	\caption{Evaluation in Mujoco. The robot wipes a slope as in Fig~\ref{fig:mujoco_slope} while tracking a sine-shaped force profile. Force measurement, mean value of predicted force and its confidence interval according to 3-$\sigma$-rule are plotted.}
	\label{fig:mujoco_plotting}
\end{figure}

The result is plotted in Fig~\ref{fig:mujoco_plotting} when a sine-wave shaped target force profile is assigned,
\begin{equation}
F(t) = 10 - 5 \sin(\frac{2\pi}{T} t),
\end{equation}
where $T$ is the time-duration of one loop of wiping motion. After training, the mean value of predicted forces and its confidence intervals according to 3-$\sigma$-rule are plotted with colored areas. It shows that in a normal situation, the learned model has a good prediction result. 


Further experiments, such as wiping over a threshold (Fig~\ref{fig:mujoco_stage}) and inside a half-circle (Fig~\ref{fig:mujoco_half_circle}) show that, without re-training the network in these scenarios, it also gives reasonable prediction results. Since the model outputs a distribution instead of a point estimate, it can still generalize to most of the situations, even if the trained model might overfit the training data,


\subsection{Result from Real System}

\begin{figure*}[ht]
		\centering
		\includegraphics[width=\linewidth]{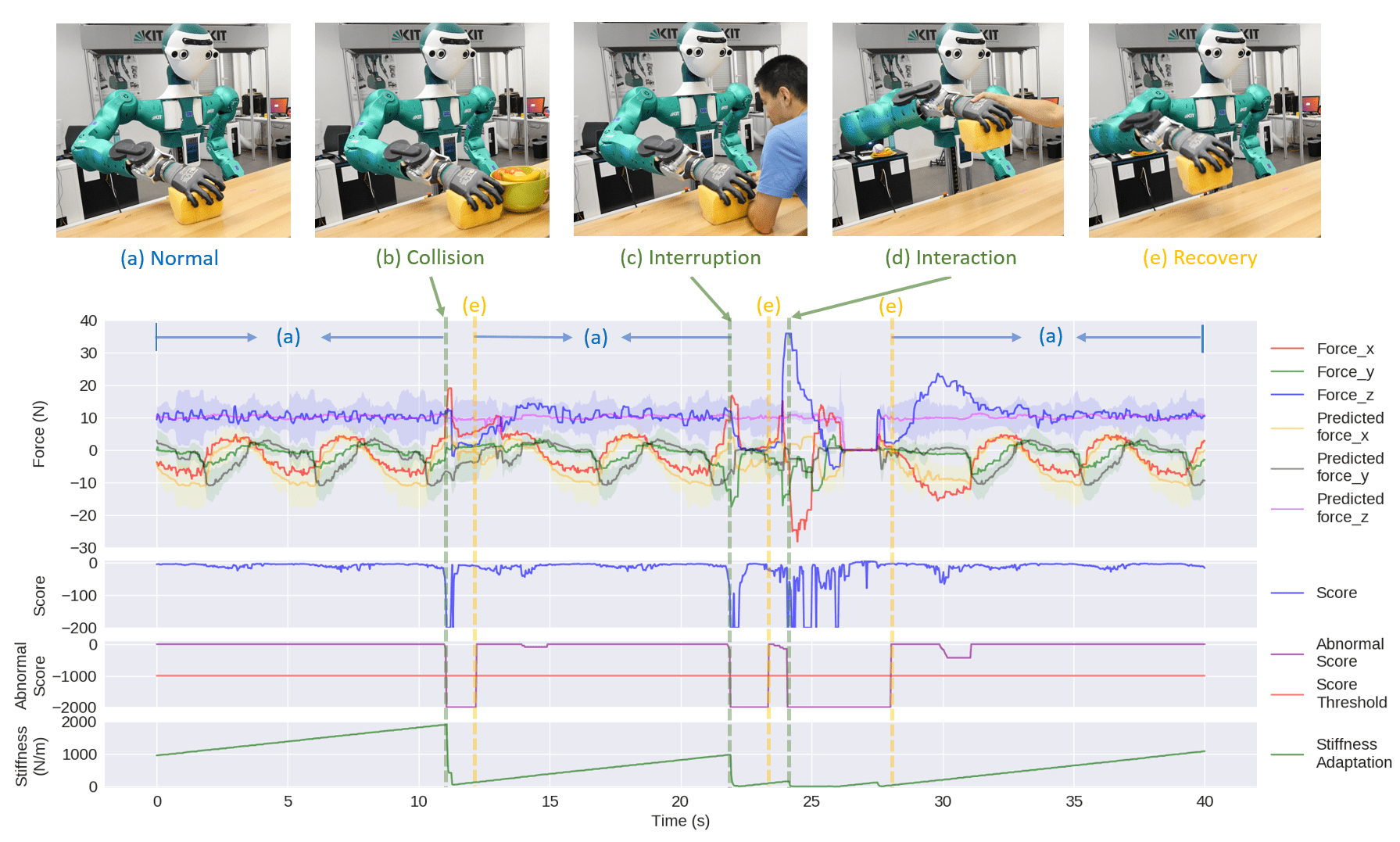}
		\caption{Force prediction and stiffness adaptation. The five photos on top of the image show the (a) normal situation, (e) recovery mode and three abnormal events (b) Collision, (c) Interruption and (d) interaction. In the lower part, the time point or period of these events are marked. The 1st row shows the actual force measurement, the mean value of predicted force and its corresponding confidence interval according to 3-$\sigma$-rule. The 2nd row shows the score values of 2 predictive models, the sum of scores and the 3rd rows shows the final abnormal score, which indicates abnormal events when it drops below $-1000$. The last row gives an example of how the stiffness of the impedance controller is adapted and recovered in different phase.}
		\label{fig:anomaly_plotting}
\end{figure*}

\begin{figure}[ht]
	\centering
	\begin{subfigure}{0.15\textwidth}
		\centering
		\includegraphics[width=\textwidth]{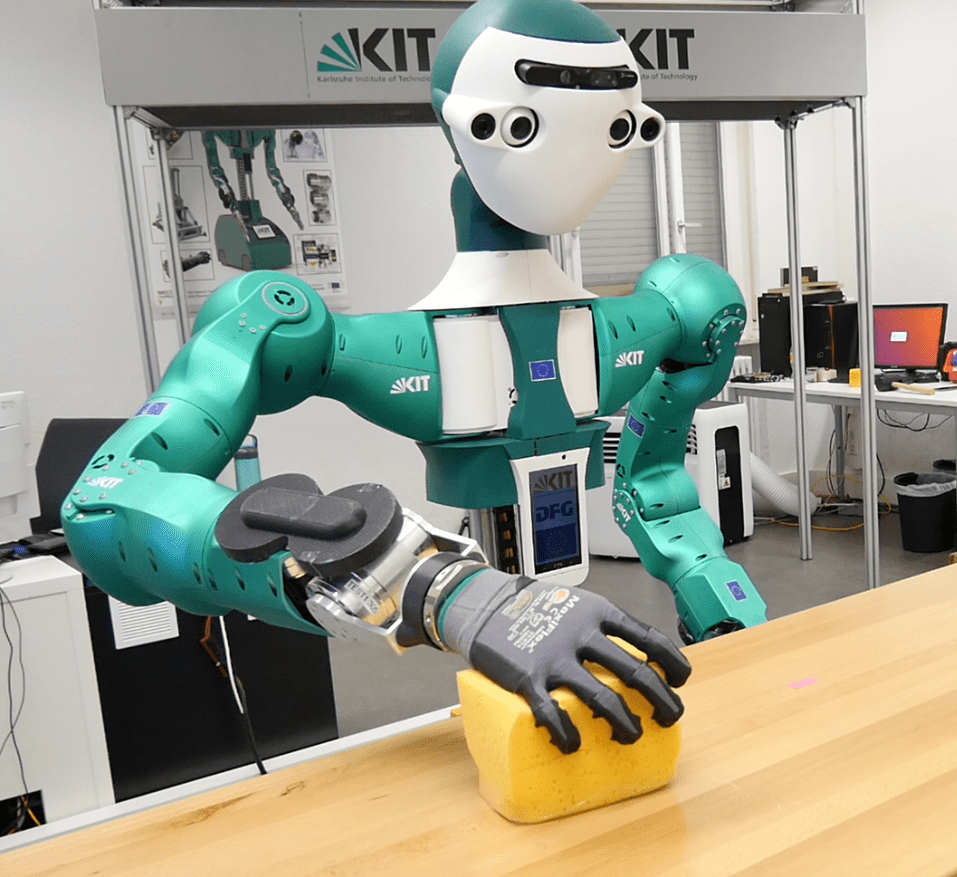} 
		\caption{Flat Surface}
		\label{fig:flat_surface}
	\end{subfigure}
	\hspace{0.00mm}
	\begin{subfigure}{0.15\textwidth}
		\centering
		\includegraphics[width=\textwidth]{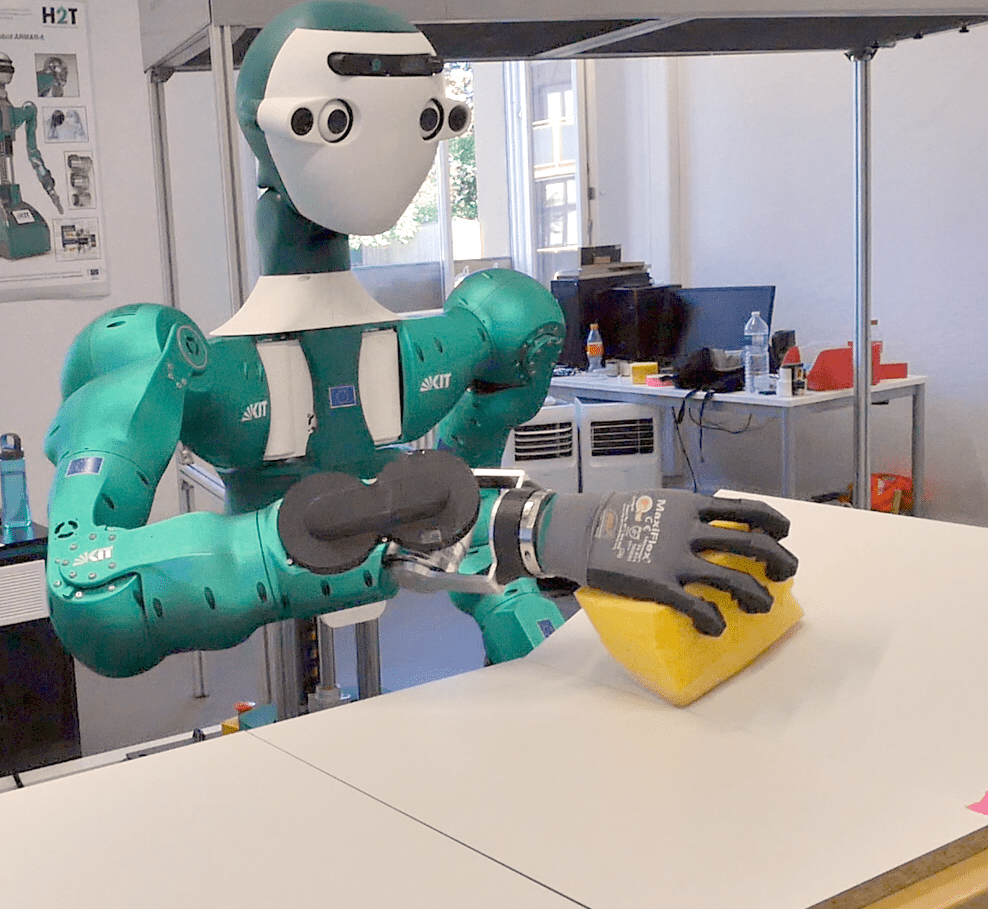} 
		\caption{Slope}
		\label{fig:slope_surface}
	\end{subfigure}
	\hspace{0.00mm}
	\begin{subfigure}{0.15\textwidth}
		\centering
		\includegraphics[width=\textwidth]{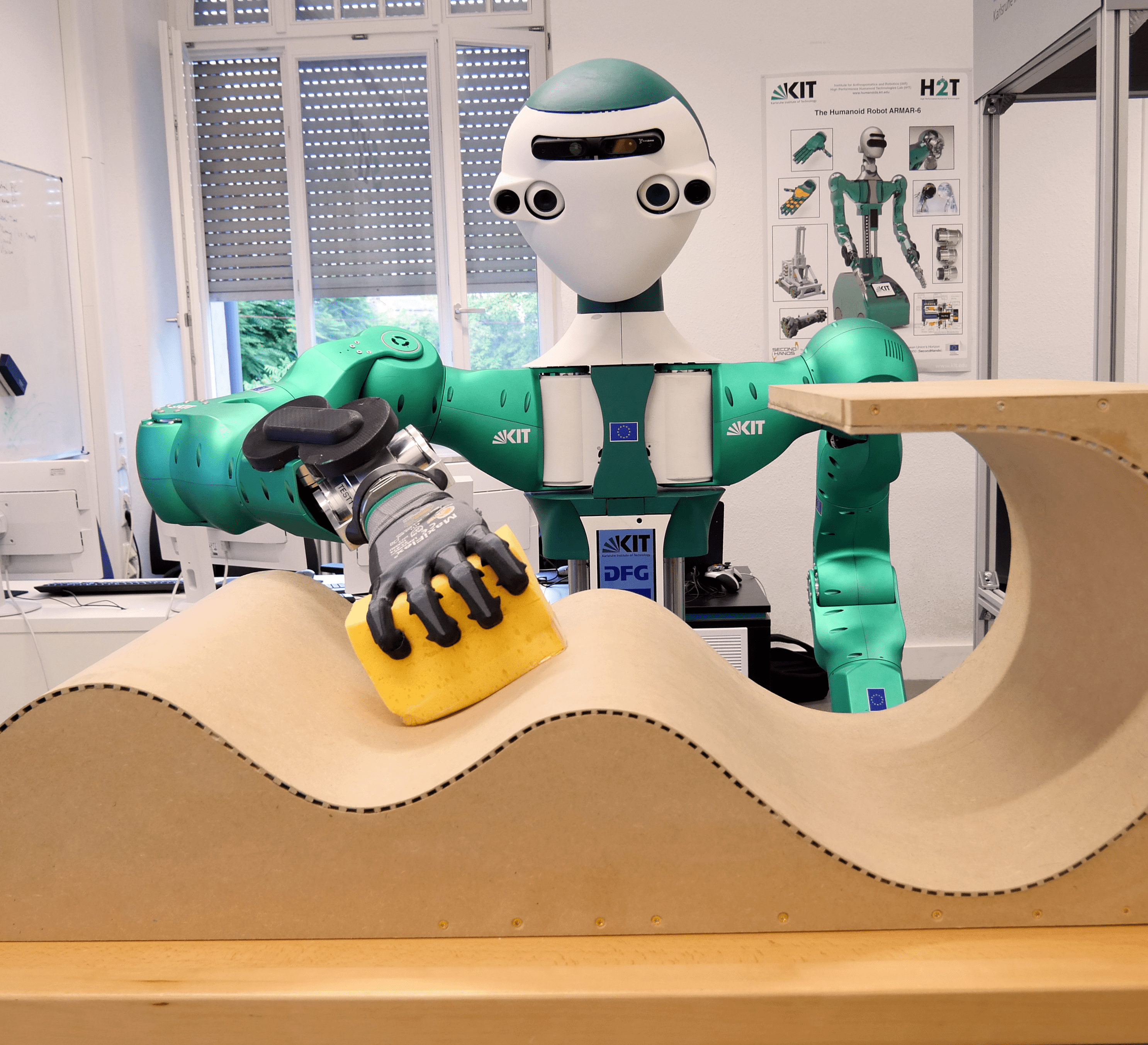} 
		\caption{Curved Surface}
		\label{fig:arbitrary_surface}
	\end{subfigure}
%
	\caption{Cleaning Task. Experiments are carried out with three different surfaces, where the robot execute the wiping task with high stiffness and keeps applying the target force perpendicular to the surface.}
	\label{fig:clean_table}
\end{figure}

For the real robot experiment, we consider the user-case scenario that the robot is wiping a table (Fig.~\ref{fig:clean_table}). 
In the normal situation, such as wiping a 
flat surface (Fig.~\ref{fig:flat_surface}), a slope (Fig.~\ref{fig:slope_surface}) and an arbitrary curved surface or a dynamic surface (Fig.~\ref{fig:arbitrary_surface}). 
The robot keeps high stiffness and high gain PID controller. The PID controllers for regulating the target force and its direction ensures that the robot applies the target force perpendicular to the surface. 
Any interference event that blocks the cleaning task is considered an abnormal situation, as shown on top of Fig.~\ref{fig:anomaly_plotting} 
(b) collision with a fruit tray, 
(c) human interruption by occupying the cleaning area,
(d) human interaction by dragging the arm away.  


The adaptive impedance controller runs on \armarVI within $1kHz$, and the force predictive model outputs the predicted force profile with $30Hz$. the force target ($10N$) is plotted in the first row of Fig~\ref{fig:anomaly_plotting}. 
In the normal situation (phase (a), corresponds to Fig~\ref{fig:flat_surface}), the root mean squared error of force regulation of the PID force controller in the target force direction ($z$ axis) is $1.047N$. 
The mean value of the predicted forces and the confidence intervals according to $3$-$\sigma$-rule are also plotted, which shows decent force tracking and prediction. 
The second and third row show the score and the abnormal score mentioned in~\autoref{subsec:adaptive_strategy}. 
With an empirical threshold, e.g. $-1000$ for wiping, the robot immediately detects the abnormal events in less than $70 ms$ as the abnormal score drops below the threshold.

The last row of Fig~\ref{fig:anomaly_plotting} gives an example of stiffness adaptation during different phases, e.g. at $11s$ the stiffness drops in $300 ms$ to $2\%$ of the value in normal case. 
Similar adaptation happens to the force and rotation PID controller,
which means these velocity sources are switched off, thus the target pose does not change much. 

After the abnormal events, the recovery mode (e) takes over and the control parameters increase back to maximum values for the high stiffness control and the normal situation is resumed.
To ensure a safe recovery, compared to the adaptive mode, the recovery of the parameter to default is set much slower to avoid control instability. In~\autoref{fig:anomaly_plotting} from $20s$ to $25s$, a recovery mode occured between two successive abnormal events, where the control parameters could jump to cause jiggled motions if their recovery was as fast as in the adaptive mode.

\section{Conclusion and Future Works} \label{sec:conclusion}

In this work, we proposed a novel adaptive compliance control by combining Bi-GRU based force predictive models with an adaptive force-impedance controller.  This controller allows the robot to 1) keep higher stiffness and high force tracking accuracy in normal situations to guarantee task execution quality and 2) be compliant when abnormal exceptions occur.   The evaluation was conducted on the \armarVI robot in the simulated and real-world environment in several use-case experiments. The results show accurate tracking behavior in the normal situation,  as well as smooth, safe recovery and compliant interaction behavior during the abnormal events.

Our approach is extensible in different manners, such as the consideration of additional velocity source and adding more force predictive models to deal with more complex tasks.  We will investigate such questions and extend the work to implement controllers, which generalizes to complex contact-rich manipulation tasks.
%
%

\bibliographystyle{IEEEtran} 
\bibliography{references}

\end{document}